\documentclass[sigconf,authorversion,nonacm]{acmart}

\usepackage{caption}
\usepackage{subcaption}

\usepackage{booktabs} 

\usepackage{array} 

\newcommand\blfootnote[1]{%
  \begingroup
  \renewcommand\thefootnote{}\footnote{#1}%
  \addtocounter{footnote}{-1}%
  \endgroup
}

\begin{document}
\title{Generating synthetic mobility data for a realistic population with RNNs to improve utility and privacy}

\author{Alex Berke, Ronan Doorley, Kent Larson, Esteban Moro 
}
\affiliation{%
  \institution{MIT Media Lab, Cambridge, MA, USA}
}
\email{{aberke, doorleyr, kll, emoro}@mit.edu}

\renewcommand{\shortauthors}{A. Berke et al.}

\begin{abstract}

Location data collected from mobile devices represent mobility behaviors at individual and societal levels. These data have important applications ranging from transportation planning to epidemic modeling. However, issues must be overcome to best serve these use cases: The data often represent a limited sample of the population and use of the data jeopardizes privacy.

To address these issues, we present and evaluate a system for generating synthetic mobility data using a deep recurrent neural network (RNN) which is trained on real location data.  The system takes a population distribution as input and generates mobility traces for a corresponding synthetic population. 

Related generative approaches have not solved the challenges of capturing both the patterns and variability in individuals' mobility behaviors over longer time periods, while also balancing the generation of realistic data with privacy. Our system leverages RNNs' ability to generate complex and novel sequences while retaining patterns from training data.
Also, the model introduces
randomness used to calibrate the variation between the synthetic and real data at the individual level.  This is to both capture variability in human mobility, and protect user privacy.

Location based services (LBS) data from more than 22,700 mobile devices were used in an experimental evaluation across utility and privacy metrics. We show the generated mobility data retain the characteristics of the real data, 
while 
varying from the real data at the individual level, 
and where this amount of variation matches the variation within the real data.

\end{abstract}

\maketitle

\blfootnote{This is the extended version of a paper submitted to The 37th ACM/SIGAPP Symposium on Applied Computing (SAC ’22). https://doi.org/10.1145/3477314.3507230}

\section{Introduction}
\label{section:introduction}

Location datasets collected from mobile devices represent the mobility behaviors of the individuals from whom they are collected. 
These data are useful for a variety of data mining and modeling applications \cite{barbosa2018human}, such as 
land use classification \cite{pei2014new},
transportation research \cite{wang2012understanding, janssens2013data},
urban planning \cite{de2016death},
epidemic modeling \cite{frias2011agent}, analyzing a pandemic \cite{andorrapaper1, googleCOVID19mobility}, 
and can serve use cases for which researchers and public agencies have typically relied on survey data~\cite{calacci2019tradeoff,ccolak2015analyzing}.

However these data have limitations  and their usage presents issues. One limitation is size. Obtained datasets often represent a small sample  of the  population, making them less useful for analyses that are meant to address the full population. Another issue is privacy. Location data can reveal sensitive information about the people whose data were collected \cite{blumberg2009locational}.

A simple approach to protect user privacy is de-identifying data by removing identifying attributes. 
However, researchers have demonstrated this is insufficient for spatiotemporal data, as knowledge of only a few points can be used to form a location-based quasi-identifier \cite{bettini2005protecting} to re-identify most users in a de-identified dataset \cite{de2013unique}. 
Prior works have attempted to mitigate these risks with strategies that modify the data,
yet researchers have shown risks are still present \cite{rossi2015spatio, murakami2016group, fiore2019privacy}. 
Moreover, these modifications decrease data utility.

This work approaches the utility-privacy tradeoff with a system to generate realistic synthetic mobility traces to be used instead of real data.
By retaining properties of the real data, the synthetic data can retain utility. 
And by sufficiently varying from the real data at the individual level, privacy risks can be mitigated.
To address the issue of limited sample sizes,
the system uses population data as input to generate synthetic data representing that population.

Our approach exploits patterns inherent in location traces by leveraging the success of recurrent neural networks (RNNs) in text generation and modeling our problem similarly.
This approach also allows
inserting calibrated randomness
 to manage variation in the model's output. This helps generate data with variation beyond the training data as well as balance the utility-privacy tradeoff.

\paragraph{Contribution.}

We present a system using an RNN to generate realistic spatiotemporal data representing individuals' mobility over extended periods.
The system takes home and work locations as inputs to generate data for a given population size and distribution.
Our work includes an experimental implementation, using a location based services (LBS) dataset, that generates data representing individuals' mobility over a 5-day workweek.
To evaluate utility we develop and use a variety of metrics that build on previous works. 
For privacy, we develop metrics to evaluate whether the variation between the synthetic and real data matches the level of variation within the real data, at the individual level.

\paragraph{Outline.} The rest of this paper is structured as follows. We discuss related works, starting with those focused on utility, and then discuss related works that address privacy, and their limitations.
We then describe our system, including
our data definitions and transformations, and how we leverage RNNs to generate synthetic location data.
We then describe our experimental implementation and evaluation, including our utility and privacy metrics and results. We conclude with a summary and future work.

\section{Related work}
\label{section:related work}

Location data from call detail records (CDRs) are similar to LBS data in that data are passively collected from mobile devices.
Many related works use 
location data from CDRs to extract information about a population's trips between places~\cite{iqbal2014development,friedrich2010generating,alexander2015origin}, as well as infer users’ home  and  work locations~\cite{hu2016home, kung2014exploring}.
Some address the problem of limited sample sizes by labeling individuals'  data with inferred home and work locations and using census data to expand datasets to match population estimates \cite{ccolak2015analyzing,jiang2016timegeo}. A common approach to such processes is to derive aggregate statistics from the real data and then use these as parameters in generative algorithms to produce synthetic trajectories for individuals. 
Many  of these generative algorithms  are designed as "Exploration and Preferential Return" (EPR) models, and are often implemented as markov chain models, where  exploration is a random walk process and  preferential return accounts for the  likelihood  of people  returning  to previously frequented  locations \cite{gonzalez2008understanding, song2010modelling, pappalardo2018data, barbosa2015effect, jiang2016timegeo}. They leverage the predictable nature of human mobility and often assume users are at predetermined home and work locations during predefined hours. 

The aforementioned works focus on data utility without addressing privacy.
Other related works use $\epsilon$-differential privacy (DP) \cite{dwork2006calibrating} in their location data publishing strategies, but without fully addressing our problem with spatiotemporal trajectory data that represents individuals over extended periods.
For example, \cite{acs2014case} considers releasing differentially private location histograms at various time intervals, but Bindschaedler et al. have noted this unsuitable for applications which require full location traces~\cite{bindschaedler2016synthesizing}.
Other related works use DP in generative algorithms. These include the n-gram model by Chen et al. \cite{chen2012Montreal,chen2012NGRAM} and the DP-Star \cite{DPSTARgursoy} and DPT \cite{DPT:he2015} frameworks. 
DPT and DP-Star generate synthetic location data with a focus on retaining spatial properties.
However, other than being time ordered sequences, data generated by DP-Star lack temporal information, and DPT is applied to trajectories that are vehicle trips rather than data observed from individuals over a broader space and time. 
Chen et al. note 
the challenge of applying DP to sequential data due to its inherent sequentiality and high-dimensionality \cite{chen2012NGRAM}. 
They use DP in generating variable length n-grams representing location trajectories. However, their approach is applied to short sequences (5 to 6 points) over a limited set of discrete locations, such as metro stations \cite{chen2012Montreal}, and \cite{DPSTARgursoy} show their approach does not retain spatial properties of data.

Noting the limitations of DP, Bindschaedler et al.  
generate spatiotemporal trajectory data to meet alternative privacy criteria, called $(k, \delta)$-plausible deniability~\cite{bindschaedler2016synthesizing}. 
Their synthetic data generation framework uses 
a  subset of seed records, (e.g. $S$), sampled from a dataset of real records, (e.g. $D$), and each seed is transformed to produce a synthetic record.
Plausible deniability requires defining a metric, $sim$, to measure similarity between trajectories, and thresholds, $\delta$ and $k$.
It can be summarized as follows\footnote{
The \emph{plausible deniability} definition is adapted from \cite{bindschaedler2016synthesizing} to contain additional context. 
}.

\emph{
A synthetic trajectory $s_i'$ generated from a seed trajectory ${s_i \in S \subset D}$ satisfies $(k, \delta)$-plausible deniability if there are at least $k \geq 1$ alternative trajectories $s_j \in D$ such that 
$|sim(s_i, s_i') - sim(s_j, s_i')| \leq \delta$.
}

Their empirical evaluation used 30 sampled trajectories and 80\% of their output satisfied their privacy criteria.

Note that the plausible deniability definition is specific to its generation framework, where each real sample seed is transformed to a synthetic record, and plausible deniability determines whether the synthetic record sufficiently differs from its seed.
However, it cannot be directly applied to more general data generation systems, such as the one presented in this work where no individual input record transforms into a synthetic output.

Another limitation of these works using DP and plausible deniability is their abstract nature.
They provide theory for how, given parameters $(\epsilon, k, \delta)$, their privacy criteria would be met. But determining parameter values, or analyzing the relationship between these values and privacy for real data, is beyond their scope.
The privacy evaluation in this work compares synthetic data to real data.

We further discuss related works in the context of how we build upon them in the following sections.

\section{Modeling the problem} 
\label{section:modeling the problem}

\begin{figure}
    \centering
    \includegraphics[width=0.45\textwidth]{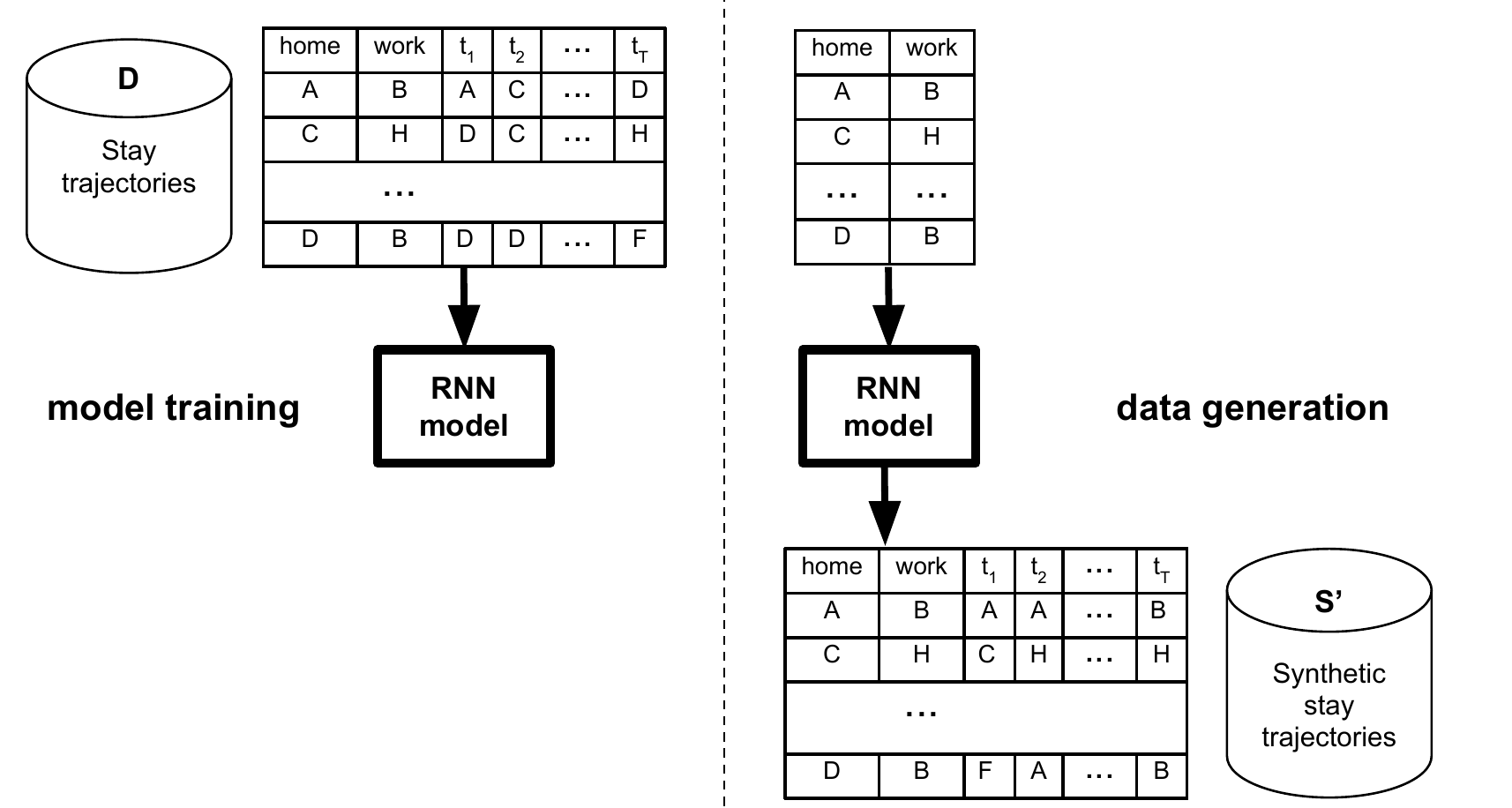}
    \caption{Model training and generation. The RNN is trained with real data, $D$, where each user’s data is a stay trajectory labeled by home, work locations.  The trained RNN takes home, work locations as input to generate synthetic data, $S'$.}
    \label{fig:rnn_model_training_vs_generation}
\end{figure}

Our system is informed by how mobility data are laden with  patterns that reflect the routines of everyday life \cite{song2010limits,schneider2013unravelling}.  Our methods exploit these patterns. 

The system's model training and generation process is summarized in Figure \ref{fig:rnn_model_training_vs_generation}. It is designed to take home and work locations as input and generate realistic location data for users with those home and work locations, based on training data, where the data represent a time period spanning multiple days or weeks.  
The system must also introduce variable so that when given the same home and work locations multiple times the generate output varies.
The generated data must also sufficiently vary from the real data in order to preserve privacy.

We use home and work locations as input because data sources such as the census provide demographic information on where people live and work.  
This information can then be used
to produce synthetic data representing realistic populations.\footnote{Even when just home locations may seem sufficient to represent populations, work locations can help generate data with sufficient variation, where people from a single home area exhibit different mobility patterns.}

\subsection{Home and Work Inference}

We define functions to infer home and work locations for use throughout this work. 
They take a single user device's location data as input and output inferred home and work locations, determined by where they spent the most time during nighttime and workday hours, respectively\footnote{What we call $work$ can be considered any secondary location to $home$.}.
These functions are used to label training data and evaluate the model output.
We also use them in our implementation with real LBS data to evaluate how well the data represents the population reported by census data (section \ref{section:experimental evaluation}).

\subsection{Data Representation}

\begin{table}[]
\small

\begin{tabular}{ |c|c|c|c|c| } 
 \hline
device ID & latitude & longitude & timestamp & dwelltime \\ 
 \hline
abc1234xyz & 42.472539 & -71.107958  & 2018-05-06-18:11:1  & 5.02 \\ 
 \hline
abc1234xyz & 42.427205 & -71.014071 & 2018-05-06-19:01:53 & 45.10 \\
 \hline
def4567qrs & 42.485207 & -71.172924 & 2018-05-07-03:17:38  & 2.03 \\ 
 \hline
\end{tabular}

\caption{Example (fake) rows from a LBS dataset.}
\label{tab:fake_LBS_table}
\end{table}

\begin{table}[]
    \small
    
\begin{tabular}{ |c|c|c|c|c|c|c|c|c| } 
 \hline
$home$ & $work$ & $t_1$ & $t_2$ & $t_3$ & $t_4$ & $t_5$ & ... & $t_{T}$ \\ 
 \hline
A  & C & A & B  & B & null & C  & ... & D \\ 
 \hline
\end{tabular}

    \caption{A "stay trajectory" prefixed by home, work locations. Locations are represented by letters.}
    \label{tab:example_stay_trajectory}
\end{table}

Location datasets from mobile devices are often timestamped geolocation coordinates, with a device ID. 
Table \ref{tab:fake_LBS_table} shows an example.
Similar to related works
\cite{pappalardo2018data, DPSTARgursoy, DPT:he2015,kulkarni2017generating},
we transform 
spatiotemporal data into sequences that discretize time and geographic space.
This results in a "stay trajectory" for each user, representing their sequence of visited locations. 
Sequence indices 
($t_1, t_2, ..., t_T$) 
represent time intervals and values are the location the user stayed for the most time within the interval.
Table \ref{tab:example_stay_trajectory} shows an example.
Locations in our "stay trajectories" are often repeated across time intervals or null valued when the device reported no data\footnote{
Instead of inferring values for missing data, this work aims to generate synthetic data with properties similar to the original dataset, including sparsity.}. 
We use census areas to represent locations, 
and map data points to their containing census areas.
We discretize the spatiotemporal data in this way so that samples have a consistent form and locations are visited a sufficient number of times for the model to learn from. Different methods may be  used to discretize the data to still work with the following methods.
We use census areas because they are
published with population estimates on how many people live and work in these geographies and are available in varying levels of granularity. 
This level of granularity and time interval size are system tunable parameters.

The values (location areas) in stay trajectories have semantic relationships. Some locations are spatially close, while others are miles apart and may be less likely to directly follow one another in a sequence.
The ideal model learns the relationships between areas and the patterns and distributions of how people spend time in them. 
It should also generate data that retain aggregate statistical properties of the real data, as well as patterns at the individual level, while also introducing variability.

We represent each stay trajectory as
$s = \langle s_1,\ s_2,\ ...,s_T\rangle$
and associate a 
$\langle home, work \rangle$ pair with each $s$, where  $home$ and $work$ are areas in $s$.









\subsection{Leveraging a Recurrent Neural Network}
\label{section:RNN}

Stay trajectories have properties similar to text and music.
Each can be represented as sequences of tokens, where they are temporal and spatial relationships between these tokens.


RNNs 
have been successful in generating complex sequences~\cite{graves2013generating} that retain the structural properties inherent in text \cite{karpathy2016unreasonable,sutskever2011generating} and music \cite{agarwala2017music,eck2002first,boulanger2012modeling}, which  we also see in stay trajectories.

RNNs are trained by processing each sequence in a training set one element at a time, and predicting a next element conditioned on  the previous elements encountered in the sequence. The loss of wrong predictions is propagated back through  the network for the model to "learn" from.  

This same process can also be used for sequence generation by feeding the model's predictions back to the model as input for the next step as if they were real data rather than the model's own inventions.  
Each prediction step samples from a distribution of candidate next elements,
where this process allows parameterizing randomness for the model's output. 
The overall process allows for the generation of novel sequences that are similar to the training set.  It also simulates a high-dimensional interpolation between training samples that distinguish RNNs from n-gram or other generative algorithms.

To leverage RNNs for our use case we prefix each $s$ with its $\langle home, work \rangle$ pair,
$\langle home, work, s_1, s_2, ..., s_T\rangle$. The prefixes serve as labels and the prefixed trajectories are used to train the  model.  
The model can then learn relationships between the prefixes and the tokens that follow, such as how tokens in the $home$, $work$ prefix positions are likely candidates for nighttime and workday hours, along with other structural relationships between tokens.

For data generation, we feed $\langle home, work \rangle$ pairs to the  trained model as labels for it to generate corresponding stay trajectories.  The RNN treats the input $\langle home, work \rangle$ pairs as prefixes for sequences it learned to complete. 
The sequences it then generates are the synthetic stay trajectories with the given  $\langle home, work \rangle$ labels.

\section{Experimental Evaluation}
\label{section:experimental evaluation}

The code for the work described in this section is open source\footnote{ 
\url{https://github.com/aberke/lbs-data/blob/master/trajectory\_synthesis}}.

\subsection{Data and Preprocessing}

\subsubsection{Data Source}
This work used LBS data provided by a location intelligence company.
The data was collected from users who opted-in to share data anonymously through a GDPR-compliant framework. Researchers followed a strict contract with obligations to not share data beyond aggregate statistics.

The data was provided as 
table rows containing a pseudo-anonymized device ID, geolocation coordinates, timestamp, and estimated time the device was in the location (see Table \ref{tab:fake_LBS_table}).

\subsubsection{Data Panel}
\label{section:data panel}

\begin{figure}
\includegraphics[width=0.33\textwidth]{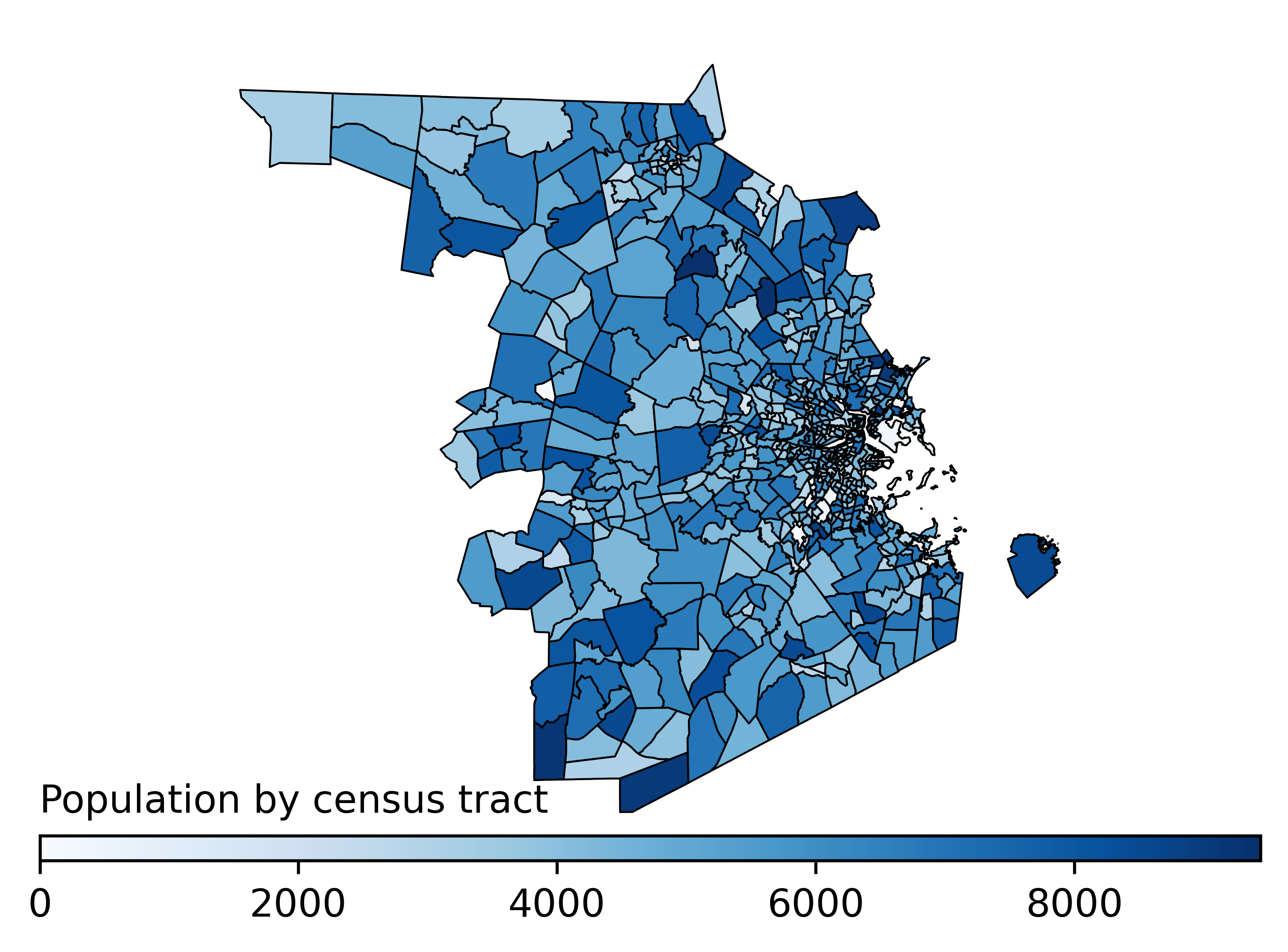}
\centering
\caption{Census tract population estimates for the geographic region represented in our dataset (ACS 2018 \cite{ACS:20185YR}).}
\label{fig:ACS2018_population_estimates}
\end{figure}

We created a panel from the first 5-day workweek of May 2018, restricted to data points reported within the 3 counties surrounding Boston, Massachusetts (Middlesex, Norfolk, and Suffolk counties). 
This region is shown in Figure \ref{fig:ACS2018_population_estimates}.
Data points representing more than 24 hours spent in one location were dropped and data was further restricted to devices reporting at least 3 unique days and 3 unique nights of data.  The resulting panel includes data from 22,707 user devices.

\subsubsection{Home and Work and Data Representativeness}

We defined functions, $inferHome$ and $inferWork$, that take a stay trajectory as input
and return census tracts for $home$, $work$.
These were used to label stay trajectories and evaluate model output.
The area the user stayed most 8pm to 9am was inferred as $home$ and the area stayed most during the remaining hours as $work$.
We applied the $inferHome$ function to the data panel to result in corresponding census tract level population estimates. Compared to ACS 2018 census estimates~\cite{ACS:20185YR} there is a Pearson correlation coefficient of 0.648.
This relatively high correlation helps validate methods and shows data representativeness\footnote{
For comparison
we used data from location data company Safegraph. 
They made statistics from their September 2019 data available, including the number of devices residing in each census area.
We measured the correlation between their device populations and census estimates at the census tract level, restricting analysis to the geographic region of our study.
This included 396,061 devices. The Pearson correlation is 0.122.
For details see 
\url{https://github.com/aberke/lbs-data/blob/master/safegraph-comparison.ipynb}.\\}.

\subsubsection{Data Granularity and Transformation}

The 5-day time period was divided into 1 hour intervals and census tracts were used as areas.  
These parameters were chosen based on panel size and data sparsity; with more data, greater spatial and temporal precision may be used. 
Stay trajectories were then created for each device in the panel with a length of 120 indices (5 days x 24 hours). There is a token vocabulary size of 652, corresponding to the study region census tracts plus the null value representing no data reported.
Stay trajectories were then prefixed with their inferred $\langle home, work\rangle$ labels\footnote{
For the purposes of
model training and data generation,
the area tokens within stay trajectories can be arbitrary. 
What is important for the model's success is the relationship between them. 
In order to publish our work in an open source repository yet keep real data private, we do the following.
We map real census areas to integers, and map areas in stay trajectories to the integers representing  the areas.
We use the transformed stay trajectories for model training  and data generation. 
The mapping between real census  areas and their integer representations is kept private.
We then map the integers in stay trajectories back to the real areas they represent when needed.
}.

\subsubsection{Data Used for Model Training, Generation and Evaluation} \mbox{} \\

\noindent
$D$: 22,707 stay  trajectories from panel. $D$ is  used to train the model. \newline

\noindent
$S$: 2000 stay trajectories randomly sampled from $D$. \newline

\noindent
$S'$: 2000 synthetic stay trajectories where the distribution of \\
$\langle home, work\rangle$ label  pairs is consistent  with $S'$.  \newline

$S'$ is generated by providing the $\langle home_i, work_i\rangle$ label pair for each $s_i \in S$ to the model.
This  results in $S'$ such that $|S| = |S'|$ and the distribution of $\langle home, work \rangle$ pairs is consistent for $S$ and $S'$.  

These consistencies are important for utility and privacy evaluation, as the evaluation compares  the real data to the synthetic data. However, any distribution or number of $\langle home, work \rangle$ pairs can then be used as input for the model to generate a synthetic dataset.

\subsection{RNN Model}

\begin{figure}
\includegraphics[width=0.23\textwidth]{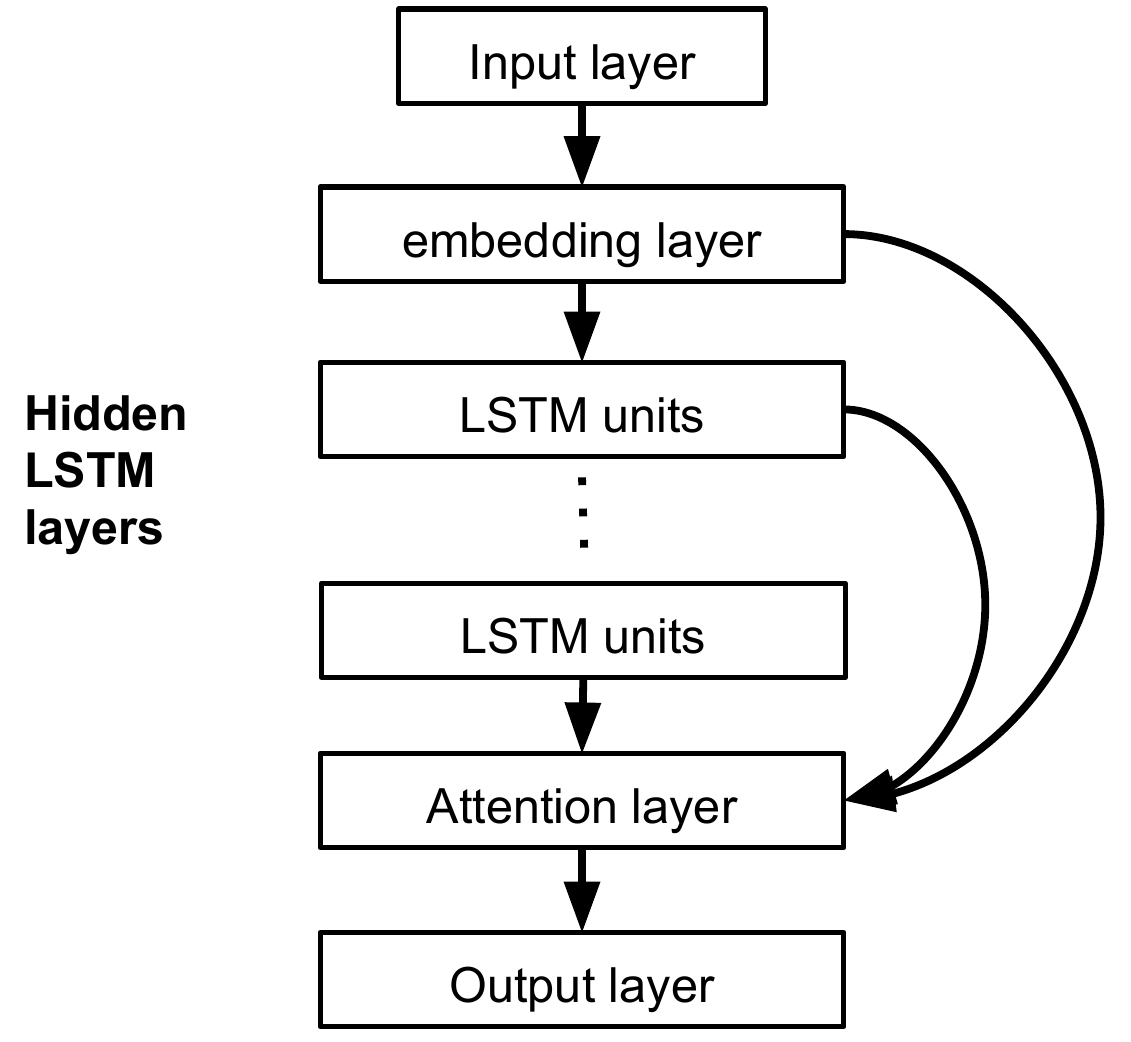}
\centering
\caption{High level RNN architecture. An embedding layer encodes input, followed by "hidden" layers of LSTM units, then an attention layer, then output layer. The embedding and LSTM layers are skip-connected to the attention layer.}
\label{fig:RNN_architecture_used}
\end{figure}

\begin{table}[]
    \small
    
\begin{tabular}{ >{\raggedright}m{2cm} | m{5.1cm} | m{0.63cm} }
 \toprule
\textbf{hyperparameter} & \textbf{description} & \textbf{value} \\
\midrule 
Embedding size & Dimension of the embedding layer. & 128 \\
\hline
Layer size & Number of LSTM units in each hidden layer. & 128 \\
\hline
Layers & Number of hidden layers. & 6 \\
\hline
Dropout & Rate at which weighted connections between neural units are randomly excluded for each training sample (a regularization method). & 0.1 \\
\hline
Maximum length & Max previous sequence tokens used to predict next token (length needed to learn patterns).  & 60 \\
\bottomrule
\end{tabular}
    \caption{RNN model hyperparameters.}
    \label{table:RNN_hyperparameters}
\end{table}

The RNN model developed in this work was intended to be simple and replicable. It was implemented via the open source textgenrnn library \cite{textgenrnn}.

The RNN is a network of neural units organized into layers with weighted connections.
The weights for the connections are learned  during the  training process, but the architecture for the network, such as the number of layers and the number of nodes within each layer, are hyperparameters that must be determined beforehand.

At a high level, the model architecture is described by Figure~\ref{fig:RNN_architecture_used} and as follows.
An exterior input layer of nodes receives input, and an exterior output layer produces the network's output.
The network learns the embedding (i.e. encoding) for input tokens via an embedding layer. 
This layer is followed by "hidden" layers of LSTM~\cite{hochreiter1997long} units, where additional layers add additional depth to a deep neural network. The LSTM layers are followed by an attention~\cite{bahdanau2014neural} layer, followed by the output layer.
The embedding and LSTM layers are each skip-connected to the attention layer, which is  connected to the final output layer.

Numerous models were trained with different hyperparameters. Their outputs were evaluated with metrics described in sections \ref{section:utility} and \ref{section:privacy} to select a best model.
These selected hyperparameters are shown in Table \ref{table:RNN_hyperparameters}.
The epoch and batch size were 50 and 1024, respectively, and a "temperature" value of 1 parameterized randomness in the predictive sampling step.
The following sections report on the output of this model.

\subsection{Utility Evaluation}
\label{section:utility}

\begin{figure*}[htbp]
     \centering
     \begin{subfigure}[t]{0.33\textwidth}
         \centering
         \includegraphics[width=\textwidth]{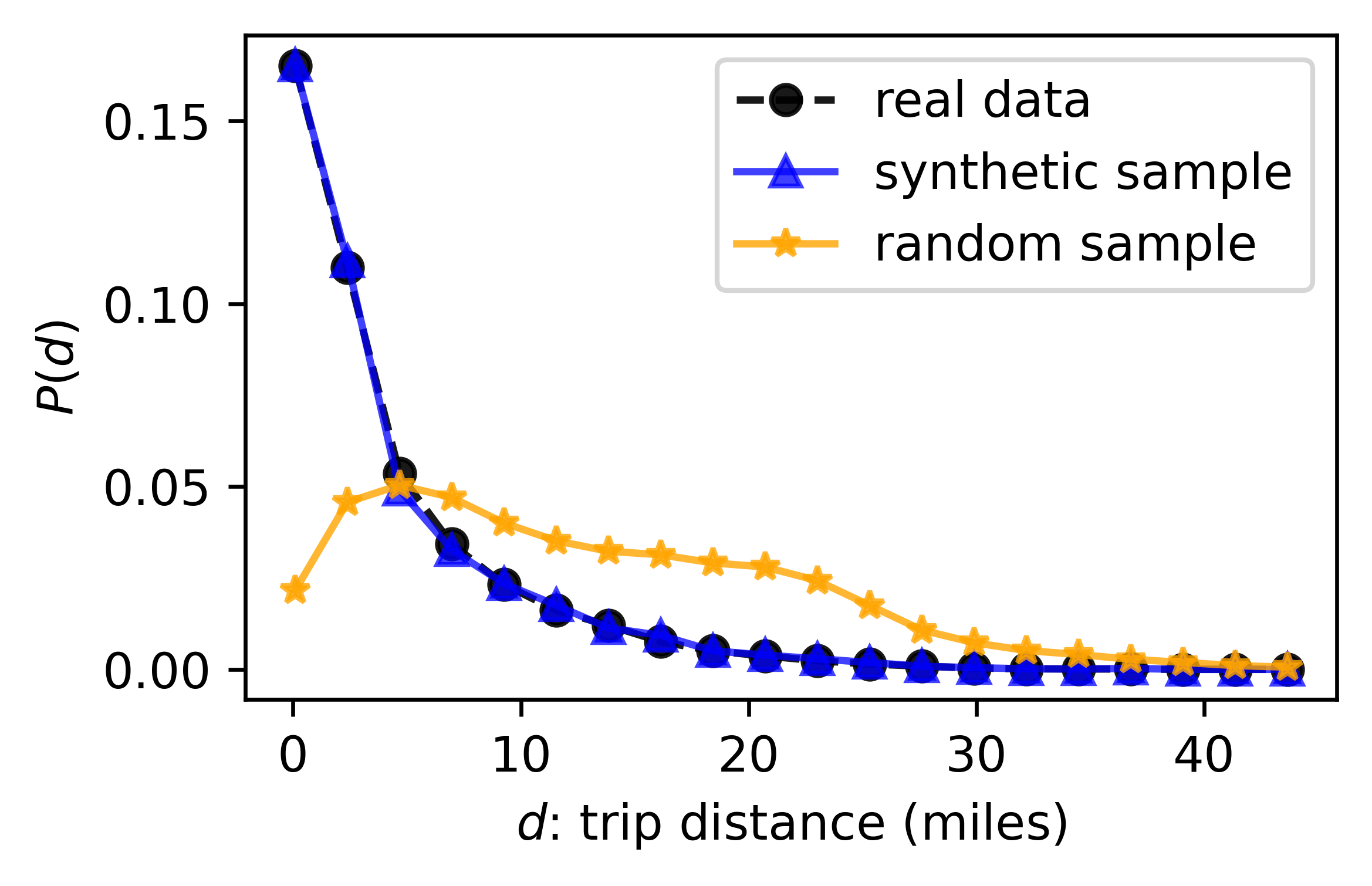}
     \end{subfigure}
     \begin{subfigure}[t]{0.33\textwidth}
         \centering
         \includegraphics[width=\textwidth]{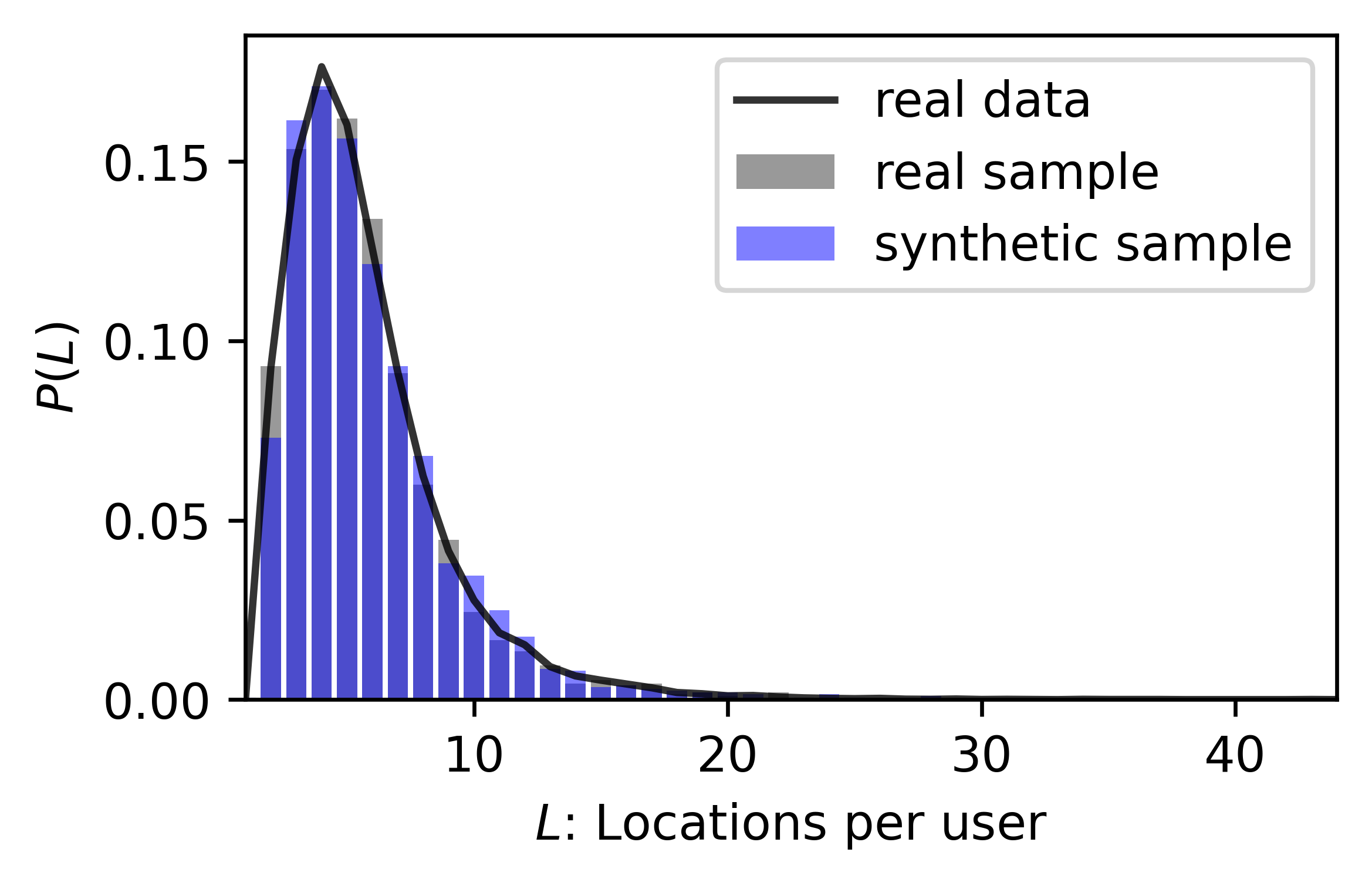}
     \end{subfigure}
     \begin{subfigure}[t]{0.33\textwidth}
         \centering
         \includegraphics[width=\textwidth]{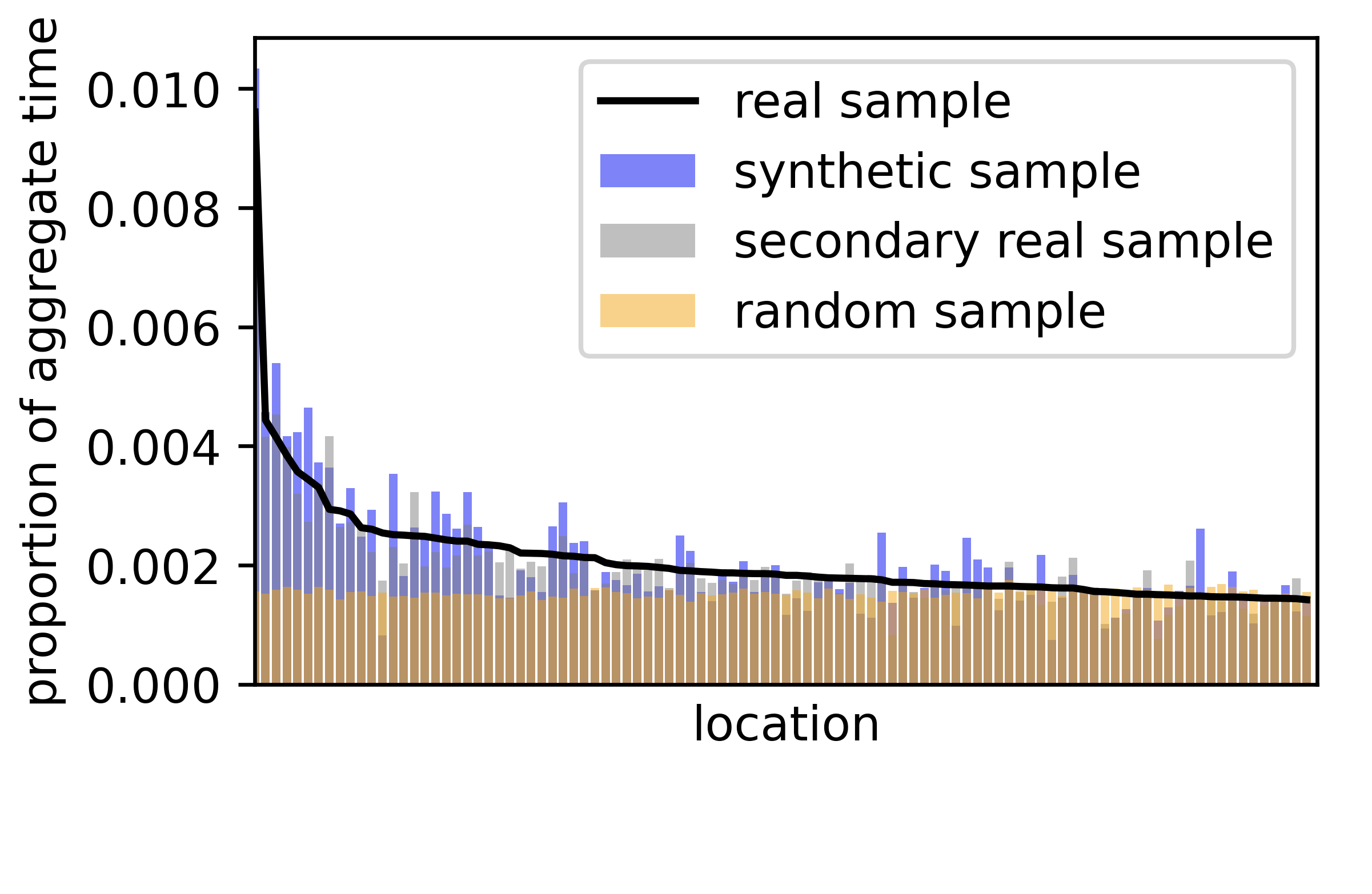}
     \end{subfigure}
        \caption{Utility metrics. (Left) Distribution of trip distances. (Center) Distribution of locations per user. The distribution for the randomly generated sample is centered beyond outliers in the real data and not shown. (Right) Proportion of aggregate time spent in each location. Locations are sorted by aggregate time for the real sample, and shown for the top-100 locations.}
        \label{fig:utility metrics}
\end{figure*}

\begin{table}[]
    \small
    
\begin{tabular}{ m{12em} m{4em} m{4em} m{5em} m{2em} }
 \toprule
\textbf{metric} & \textbf{synthetic sample} & \textbf{real\newline secondary\newline  sample} & \textbf{randomly\newline generated\newline sample} \\

\midrule 
trip distance \\(KL divergence) & 0.0008 & 0.0015 & 0.3655 \\
\hline
locations per user \\(KL divergence) & 0.0124 & 0.0044 & -2.4587 \\
\hline
aggregate time per location \\(KL divergence) & 0.0366 & 0.0085 & 0.9608 \\
\hline
home label error rate & 0.1375 & 0.0863* & 0.9995 \\
\hline
work label error rate & 0.2675 & 0.2415* & 0.9235 \\
\bottomrule
\end{tabular}
    \caption{Utility metrics evaluating error from real data. Two baselines are used for comparison: A secondary real data sample and randomly generated sample. Lower values indicate lower error. *Measured as the rate at which labels change between weeks (section \ref{section:utility:home_work_error}).}
\label{table:metrics_results}
\end{table}

Related works evaluate utility by the extent to which synthetic data retains the characteristics of real data.  
Evaluations have used a variety of quantitative metrics as well as visual mechanisms. (For example, \cite{DPSTARgursoy} plot spatial densities of synthetic taxi trip origins and destinations. 
Similarly, we visualize mobility patterns at the individual level by plotting stay trajectories and compare plots between the real and synthetic samples. See Appendix \ref{appendix:mobility plots}.)

We build on metrics from previous works \cite{pappalardo2018data, jiang2016timegeo, bindschaedler2016synthesizing}
and borrow their evaluation strategy of using Kullback-Leibler divergence \cite{cover1999elements} to compute differences between metric distributions for the real versus synthetic data.
We also evaluate how well synthetic trajectories match their input labels.
Results are shown in Table~\ref{table:metrics_results} and Figure~\ref{fig:utility metrics}. 

Note that even when using evaluation methods from related works, we cannot directly compare the output metrics between works because the evaluations are for different data generation processes developed around different goals and are done over different datasets. Indeed, even studies that apply the same generation processes and evaluation methods to multiple datasets yield different results for each dataset \cite{DPSTARgursoy, pappalardo2018data}.

Many related works use utility metrics to compare versions of their own data generation processes, such as various trip generation or privacy parameters \cite{pappalardo2018data, chen2012NGRAM, DPSTARgursoy, DPWHERE:mir2013dp}.
We did this as well to select our best model parameters.
To evaluate overall utility, baselines are needed. 
To this end, Bindschaedler et al. \cite{bindschaedler2016synthesizing} draw 2 baseline samples: (1) a second sample of real data (where synthetic data was generated from the first sample) and (2) uniformly random location traces.  We follow their approach to draw 2 baseline samples as well, where each sample matches in size (|$S$|=2000) and $\langle home, work \rangle$ label distribution.  

\paragraph{Secondary real sample:} Randomly drawn with replacement from $D$ s.t. the distribution of $\langle home, work \rangle$ labels is consistent with $S$.

\paragraph{Randomly generated sample:} For each $\langle home, work \rangle$ label, each area in a stay trajectory is drawn from a uniform distribution over the vocabulary of tokens.

Similar utility metrics between the synthetic and secondary real samples may imply that the synthetic generation process is similar to drawing from the real dataset in terms of utility.

\subsubsection{Trip Distances}

Previous studies have characterized human mobility by distributions of travelling distances \cite{brockmann2006scaling, gonzalez2008understanding} and a class of generative algorithms aims to realistically reproduce these distributions \cite{pappalardo2018data, jiang2016timegeo}. 
We borrow their methodology as follows.
Two consecutive locations in a stay trajectory are considered a trip when the locations differ (and neither is null valued). Each location is a census area; shapefiles are used to measure trip distance as the geographic line distance between the centroids of the areas.
For each sample and the real dataset, $D$, distance distributions are transformed into discrete probability distributions, $P(d)$, via a histogram (20 bins).
The distributions are plotted and the sample distributions are compared to the real distribution ($D$) with KL divergence.
See Table \ref{table:metrics_results} and Figure \ref{fig:utility metrics}. 

The distributions for the synthetic sample and real data closely match.
This match is even closer than that for the secondary real sample, which we attribute to the randomness of sampling from the real data.
In contrast, the distribution for the randomly generated sample is an entirely different shape, demonstrating that the distribution for the synthetic data is not simply a result of the distribution of distances between areas.\footnote{Note that in our dataset, all trips could have been realistic in terms of distance - the largest distance spanned between centroids of geographic areas was $<$60 miles, which could be feasibly traveled by car within the 2 hours corresponding to 2 intervals.}


\subsubsection{Locations Per User}


This metric is used in generative "Exploration and Preferential Return" (EPR) models to describe the degree of "exploration" of individuals, as well as evaluate synthetic data \cite{jiang2016timegeo, pappalardo2018data}. 
Following their methods, $L$ is locations per user and we compute the distribution of locations per user, $P(L)$. KL divergence computed across $P(L)$ is then used to evaluate the synthetic sample against $D$.
See Table \ref{table:metrics_results} and Figure \ref{fig:utility metrics}.

We observe large variation in locations per user in $D$. Some stay trajectories have only 2 or 3 distinct locations, while others have tens.
This variation may be due to heterogeneity in users’ levels of activity or activity diversities, as well as some devices reporting data more often. 
This variation and distribution should be maintained.

The distribution for $S'$ closely matches $D$. In contrast, the distribution for the randomly generated sample is centered well beyond outliers for $D$.

We then take this evaluation further than related works and use Pearson's chi-squared test for homogeneity \cite{pearson1900x}. 
In particular, we consider $S'$ as a sample that might be drawn from the real population, $D$, and test whether the Pearson's chi-squared test would discern otherwise, with respect to the distribution of locations per user.

Locations per user are binned into 6 equal quantiles determined by the distribution in $D$. Each bin is used as a category in the test, and expected frequencies are computed from the proportions of frequencies in $D$. 
We test the following null hypothesis with a significance level of 0.05:
The proportion of $s'$ in $S'$ with $l$ locations is the same as the proportion of $s$ in $D$ with $l$ locations.

For comparison, testing the real data sample, $S$, against $D$, results in a p-value of 0.841.
Testing $S'$ results in a p-value of 0.429, allowing us to keep the null hypothesis that the distributions are consistent between the synthetic and real data.

\subsubsection{Proportion of Aggregate Time Spent Per Location}

Individual stay trajectories should vary in where the users they represent go, and when. 
Yet, in aggregate the distributions of where users spend time should be consistent across the real and synthetic data. 
We compute the aggregate time intervals spent in each area for the real data sample, $S$, and synthetic and baseline samples, and measure both the Pearson correlation and KL divergence between their distributions. 
Comparison is made to $S$ rather than $D$ because the distribution of $\langle home, work \rangle$ pairs is then consistent and where users spend time is biased to where they live and work. 

To illustrate results, we plot the proportions of aggregate time spent for each sample for the top-100 locations (Figure \ref{fig:utility metrics}). The correlations and KL divergence are computed over all possible 651 locations.
The Pearson correlation coefficient between the distribution of aggregate time spent in each area, when comparing $S$ and $S'$ is $\rho=0.9366$ (p=0.0000). The correlations for the secondary real sample and random sample are 0.9652 (p=0.0000) and -0.0280 (p=0.4755), respectively.


\subsubsection{Home and Work Label Error Rate}
\label{section:utility:home_work_error}

The model is  designed to take
$\langle home, work\rangle$ pairs as input and output corresponding synthetic stay trajectories representing users with those home, work locations.
For each input label $\langle home_i, work_i\rangle$ and output $s_i' \in S'$,
we count errors when $home_i \neq inferHome(s_i')$ and $work_i \neq inferWork(s_i')$,
and calculate error rate as total errors over |S'|.

What is a reasonable error rate? Variation exists within the real data which we use as a baseline.
We draw a secondary data panel using the same criteria used to produce $D$ (section \ref{section:data panel}) but for the following 5-day workweek. 62\% of user devices in the first panel are also in the second panel. 
Error for real data is then calculated as the rate inferred labels changed for users between weeks.
Home and work label error rates for the real data are 0.0863 and 0.2415, respectively, compared to 0.1375 and 0.2675 for $S'$ (Table \ref{table:metrics_results}).

\subsection{Privacy Evaluation}
\label{section:privacy}

Well established privacy frameworks do not readily apply to the synthetic spatiotemporal trajectory data in this work.
For example, $k$-anonymity \cite{sweeney2002k}, which has been adapted and widely used for spatiotemporal data \cite{fiore2019privacy}, addresses the risk of users being re-identified in a de-identified database. 
However, in a synthetic database, there are not real users to be re-identified.
Even so, privacy risks remain if synthetic records are too similar to real records.
Section \ref{section:related work} notes the challenges of applying DP to spatiotemporal trajectory data, and how related works have therefore limited their application of DP to different problems. 
Bindschaedler et al. developed \emph{plausible deniability} to more directly address synthetic trajectory data \cite{bindschaedler2016synthesizing}. 
A central concept to both DP and plausible deniability is to make it difficult to distinguish whether an individual's record was included in a dataset or sample.
Plausible deniability approaches this by using a metric to check that a synthetic trajectory is not too similar to a corresponding real seed trajectory, 
when compared to other real trajectories.
We build on these concepts when considering the synthetic data, $S'$, as an alternative to $S$, which was randomly sampled from the real dataset, $D$.
Our evaluation measures 
the similarity between any synthetic trajectory in $S'$ and any real trajectory in $D$, and
checks that trajectories are not too similar, when compared to the similarities between real sample trajectories in $S$ and any other real trajectories in $D$.
In addition, the model should generate varying synthetic data even when given the same input, which is also evaluated.

\subsubsection{Privacy Evaluation: Method}

\paragraph{Distance metric.}

For any two trajectories, $s_i$, $s_j$, we measure the difference between them, $d(s_i, s_j)$, as the Levenshtein edit distance, which is a metric developed for sequences \cite{levenshtein1966binary}. It measures the minimum number of insertions, deletions, or substitutions necessary to transform one sequence into the other.

Related works have used various metrics for location trajectory data, such as Euclidean distance \cite{fu2005similarity},
Longest Common Subsequence \cite{buzan2004extraction}, 
Hausdorff distance \cite{junejo2004multi,lou2002semantic}, Manhattan norm \cite{torres2016frechet}, and compared them \cite{zhang2006comparison,EDR:chen2005robust}. 
\cite{EDR:chen2005robust} found edit distance\footnote{Chen et al. (2005), refer to the edit distance used in their paper as "Edit Distance on Real sequence" (EDR) \cite{EDR:chen2005robust}. It is based on Levenshtein's edit distance and modified to handle real-valued locations, as opposed to the discrete values.}
a more robust metric in terms of accuracy and accounting for noise, and we find it most applicable to our use case.




\paragraph{Comparing minimum distances.}

We compute the minimum distance between a given $s$ and any other $s_j$ in $D$, which we call ${min\text{-}dist(s, D)}$.  
\begin{displaymath}
    min\text{-}dist(s, D) = d(s, s_j)\ s.t. \forall\ s_j, s_k\ \epsilon\ D, d(s, s_j) \leq d(s, s_k)
\end{displaymath}

These values are computed for each $s \in S \subset D$, but where direct comparison of $s$ to itself is avoided.

Similarly, we compute for each $s' \in S'$,
\begin{displaymath}
    min\text{-}dist(s', D) = d(s', s_j)\ s.t. \forall\ s_j, s_k\ \epsilon\ D, d(s', s_j) \leq d(s', s_k)
\end{displaymath}

We also use the same model that generated $S'$ to generate another synthetic sample, $S''$, over the same distribution of $\langle home, work \rangle$ label pairs used to generate $S'$.
This is to evaluate whether the model generates data with sufficient variation, given the same input.
To compare $S''$ to $S'$, we compute for each $s'' \in S''$,
\begin{displaymath}
    min\text{-}dist(s'',S') = d(s'', s_j')\ s.t. \forall\ s_j', s_k'\ \epsilon\ S', d(s'', s_j') \leq d(s'', s_k')
\end{displaymath}

Our evaluation considers for a distance $m$, and probability $\delta$, 
\begin{displaymath}
    Pr[ min\text{-}dist(s, D) \leq m ] \leq \delta
\end{displaymath}

And we evaluate privacy criteria for any distance $m$,
\begin{displaymath}
Pr[ min\text{-}dist(s', D) \leq m ] \leq  Pr[ min\text{-}dist(s, D) \leq m ]
\end{displaymath}

Meaning that the probability that a synthetic trajectory, $s'$, differs from any real trajectory in $D$ by less than $m$, is less than or equal to the probability that a real sampled trajectory, $s$, differs from other real trajectories in $D$ by less than $m$.

Probabilities describe the evaluation since sampling $S$ and generating $S'$ are stochastic processes.
For our experiment, we evaluate empirical distributions.
i.e.
${Pr[ min\text{-}dist(s, D) \leq m]}$ is estimated as the proportion of $min\text{-}dist(s, D)$ values where $min\text{-}dist(s, D) \leq m$.
The same is done for $S'$ and $S''$.

To compare the distributions over the range of $m, \delta$ values, we can use Q-Q plots.
We also use benchmark values of $\delta$ to check 
the maximum values of $m$ that satisfy ${Pr[ min\text{-}dist(s, D) \leq m ] \leq \delta}$, and compare these values of m across the distributions for $S, S', S''$.
For each value of $\delta$, a higher corresponding value of $m$ indicates a greater minimum distance from $D$. i.e. higher corresponding values of $m$ are desired for the $min\text{-}dist(s', D)$ and $min\text{-}dist(s'', S')$ distributions for more privacy.
Benchmark values of $\delta$ were used to compare model outputs when selecting the best model parameters.

Since $S \subset D$, values for $min\text{-}dist(s, D)$ are computed as\\ ${min\text{-}dist(s, D \setminus S)}$ to avoid yielding a distribution of 0's.\footnote{
The $s$ in $S$ are removed from $D$ only once; any $s$ with a duplicate in $D$ will still be found in $D\setminus S$, resulting in a $min\text{-}dist$ value of 0. 
This  does  occur in our dataset.} 
The values are still referred to as $min\text{-}dist(s, D)$ for notational convenience and $min\text{-}dist(s', D)$ values are still computed over all of $D$.
When comparing the $min\text{-}dist(s, D)$ and $min\text{-}dist(s', D)$ distributions, $s'$ and $s$ with unique $\langle home, work \rangle$ pairs are removed.
This makes for a more balanced comparison between $min\text{-}dist$ distributions for $S$ and $S'$ since 
any $s \in S$ with a unique $\langle home, work \rangle$ pair already avoids comparison to any $s \in D$ with the same pair when $min\text{-}dist(s, D \setminus S)$ is computed.
It also better preserves privacy\footnote{Research from 2009 found that  more than 5\% of individuals in the U.S. working population have unique combinations of home and work census tract locations \cite{golle2009anonymity}.} and is line with the method of \cite{bindschaedler2016synthesizing}
where synthetic trajectories that do not sufficiently differ from enough real trajectories are discarded. 
When comparing the distribution of $min\text{-}dist(s'', S')$, all stay trajectories are used.

\subsubsection{Privacy Evaluation: Empirical Results}

\begin{table}[]
    \small
    \centering
    
\begin{tabular}{ p{1.5cm} p{1.75cm} p{1.8cm} p{1.8cm}}
 \toprule
 & \textbf{$min\text{-}dist(s, D)$} &  \textbf{$min\text{-}dist(s', D)$} & \textbf{$min\text{-}dist(s'', S')$}
\\

\midrule
minimum &  0 & 2 & 1 \\
$\delta=0.01$  & 5 & 5 & 8 \\
$\delta=0.05$  & 11 & 10 & 15 \\
$\delta=0.10$ & 14 & 14 & 20 \\
$\delta=0.25$ & 22 & 22 & 30 \\
\bottomrule
\end{tabular}

    \caption{Privacy evaluation empirical results: distributions of minimum distances compared between samples.}
    \label{table:min_dist_values}
\end{table}

\begin{figure}
     \centering
     \begin{subfigure}[b]{0.27\textwidth}
         \centering
         \includegraphics[width=\textwidth]{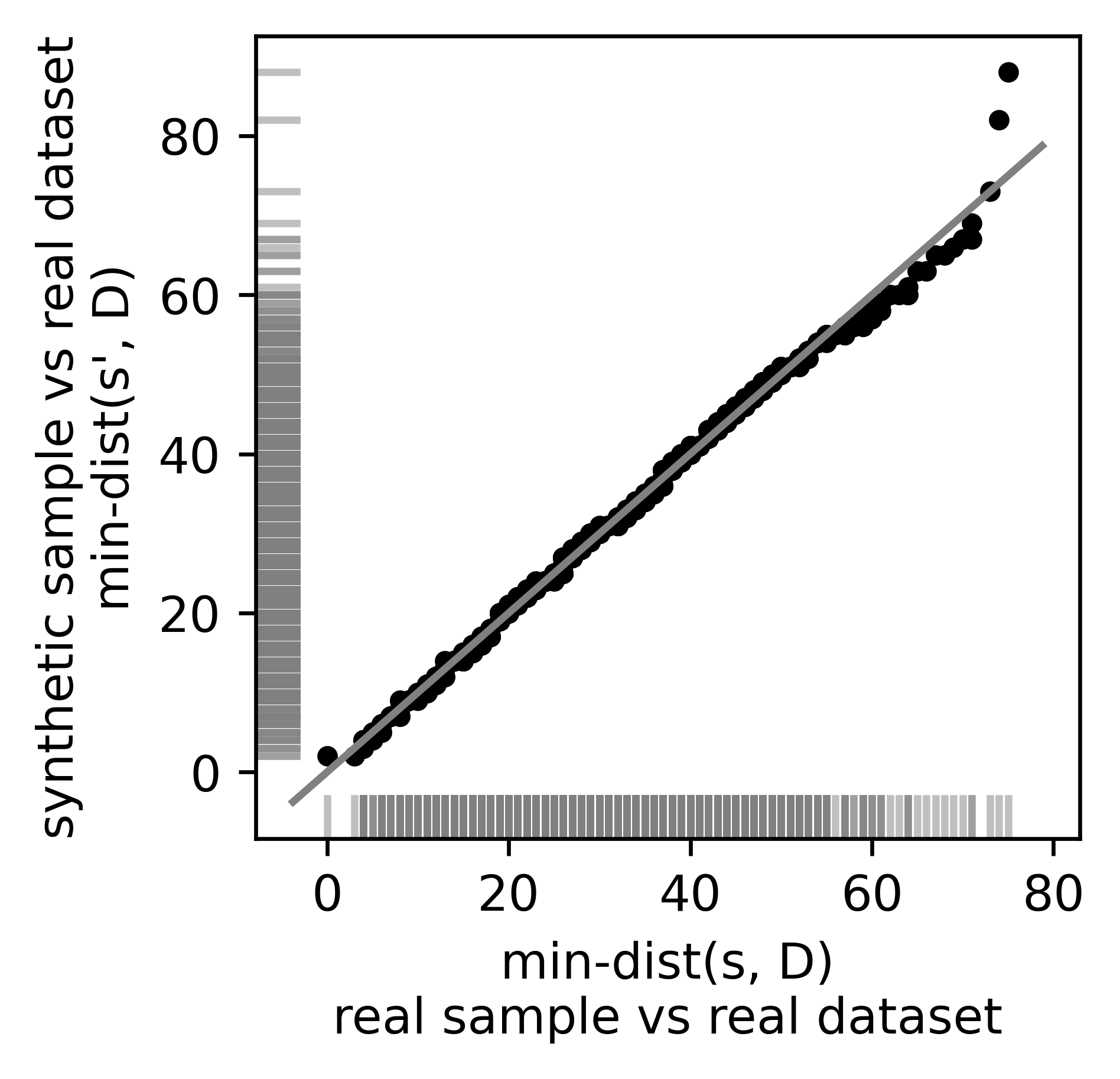}
     \end{subfigure}
     \begin{subfigure}[b]{0.27\textwidth}
         \centering
         \includegraphics[width=\textwidth]{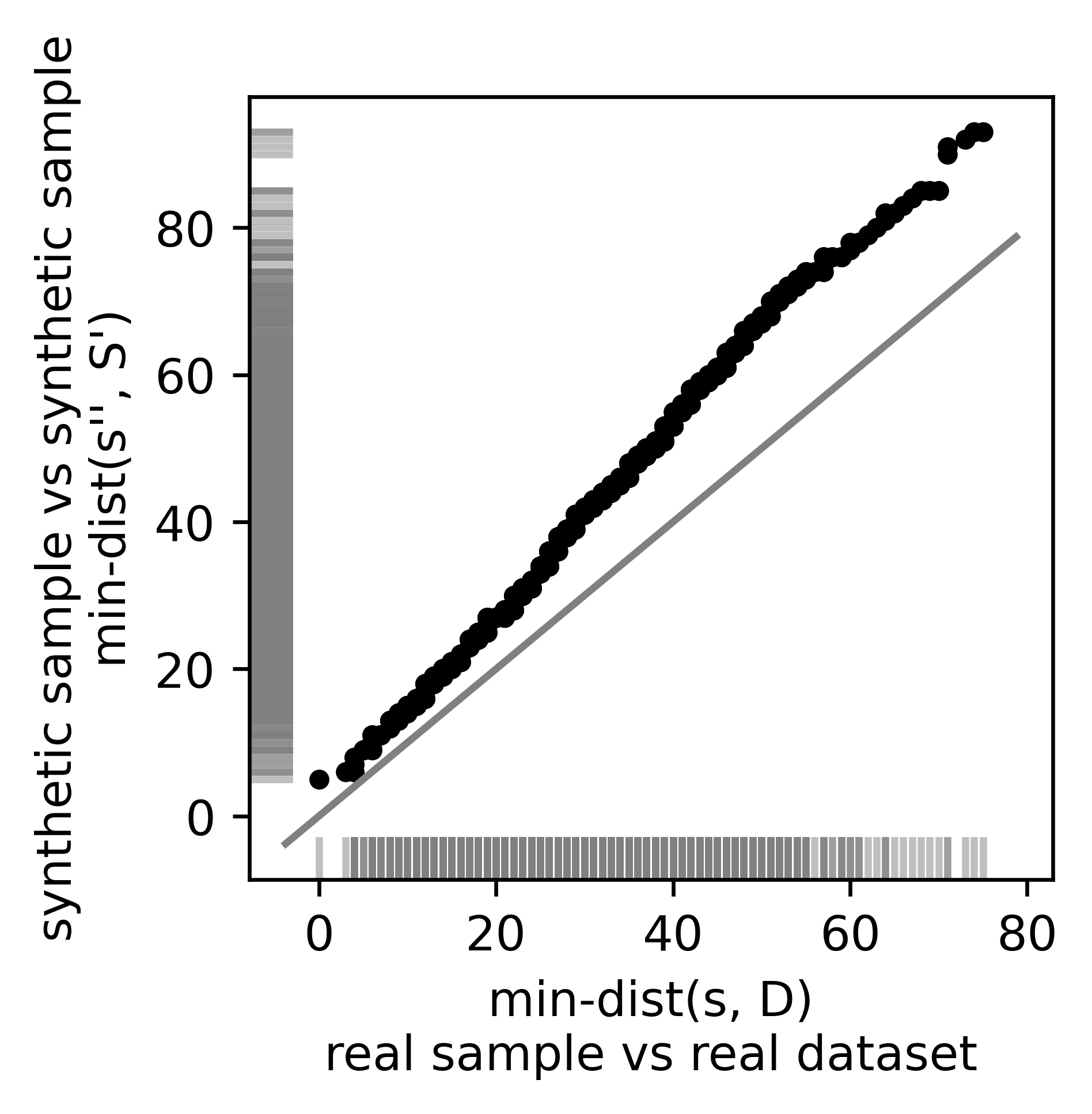}
     \end{subfigure}
        \caption{Q-Q plots comparing $min\text{-}dist$ values.}
        \label{fig:qq_plots}
\end{figure}

See Figure \ref{fig:qq_plots} (top) for the Q-Q plot evaluating $S'$.
The distributions of $min\text{-}dist$ values are sorted and the  Q-Q  plot matches the corresponding  $m$ values for the $S$ and  $S'$  against  each other,  with values for $S$ and $S'$ on the x  and y axes,  respectively. 
Each  point then shows data  for a different  $\delta$, where the point's  x-value is  the corresponding $m$  value for $S$, and the point's y-value is  the  corresponding $m$ value for $S'$.  A 45-degree line represents a matching distribution of values, and  points on or above  the  45-degree  line represent  where  the privacy   criterion  is met. 
The values closer to the origin are more important, as these are for smaller $min\text{-}dist$ values, $m$, where privacy risk is higher.

Our experiment results show the comparison of $min\text{-}dist(s', D)$ to $min\text{-}dist(s, D)$ closely tracks the 45-degree line, without perfectly satisfying the privacy criterion.
Similarly, the distribution of $min\text{-}dist(s'', S')$ is compared to the distribution of $min\text{-}dist(s, D)$ in order to evaluate whether the model can generate multiple samples that sufficiently vary from one another, when given the same input. 
Figure \ref{fig:qq_plots} (bottom) shows points on the Q-Q plot above the 45-degree line, showing that the difference between the synthetic samples, $S''$ and $S'$, tends to be greater than the difference within the real data when $S$ and $D$ are compared.

Benchmarks are also established with $\delta$ values of 0.01, 0.05, and 0.10 to compare corresponding cutoff $m$ values for the $min\text{-}dist(s, D)$, $min\text{-}dist(s', D)$, $min\text{-}dist(s'', S')$ distributions. 
Results are shown in Table \ref{table:min_dist_values}.
Overall, these results imply the model generates synthetic data that nearly meets the level of variation within the real data, at the individual level.

\section{Conclusion and Future Work}

This work presented a system to use real data to generate synthetic location trajectory data, with methods that exploit patterns within human mobility behaviors, using the mechanisms of RNNs.

A synthetic data sample was presented and evaluated as an alternative to a real data sample, 
to offer utility in data mining and modeling applications, while mitigating privacy risks since it represents synthetic individuals.
To this end, a variety of utility metrics evaluated how well the synthetic data retained the properties of the real data. 
The privacy evaluation measured whether the level of variation between the synthetic and real data matched the level of variation already present between the real user data.

For the purposes of evaluation, a synthetic data sample with home, work labels matching the distribution of a real sample was used. 
Yet the system can then be used to generate data for another given population distribution, such as that matching census data or for modeling alternative populations. 
Since the model generates data with sufficient variation, it can then be used to generate data for much larger synthetic populations.

Our experiment used a LBS dataset with methods that future work can extend and test using other forms of location data or applications. 
Furthermore, related works have approached the challenge of merging location data from various sources to de-duplicate data from the same user \cite{wang2019extracting}. 
In contrast, our machine learning approach may offer a new opportunity to combine various data sources in model training. 
This may improve synthetic data generation via more training data while avoiding the issue of de-duplicating data. 
Future work can test this as well.


\bibliographystyle{ACM-Reference-Format}
\bibliography{main} 

\appendix

\clearpage 

\section{Mobility Pattern Plots}
\label{appendix:mobility plots}

\begin{figure}
     \centering
     \hspace*{-0.65cm}
     \begin{subfigure}[b]{0.55\textwidth}
         \centering
         \includegraphics[width=\textwidth]{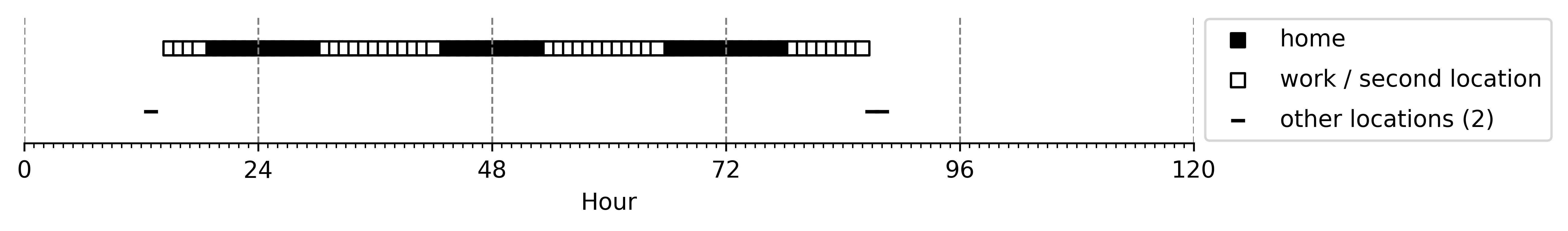}
     \end{subfigure}
     \hspace*{-0.65cm}
     \begin{subfigure}[b]{0.55\textwidth}
         \centering
         \includegraphics[width=\textwidth]{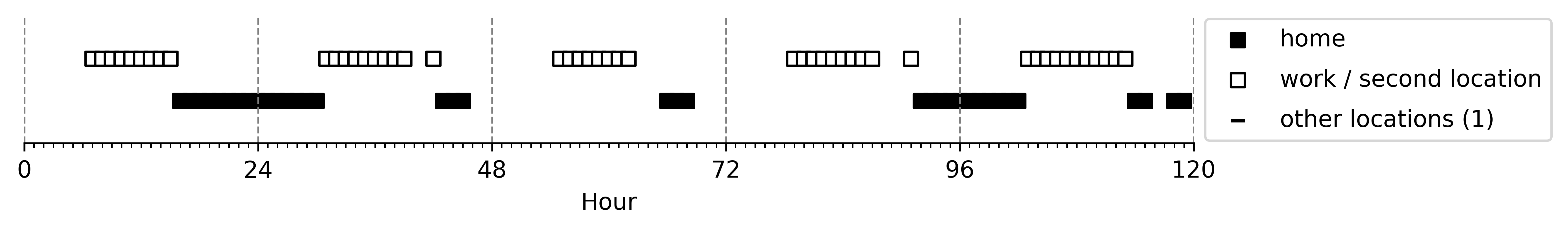}
     \end{subfigure}
     \hspace*{-0.65cm}
     \begin{subfigure}[b]{0.55\textwidth}
         \centering
         \includegraphics[width=\textwidth]{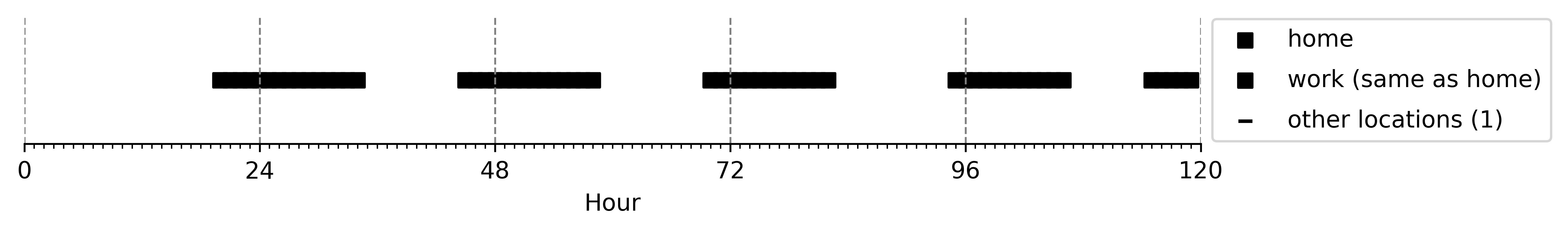}
     \end{subfigure}
     \hspace*{-0.65cm}
     \begin{subfigure}[b]{0.55\textwidth}
         \centering
         \includegraphics[width=\textwidth]{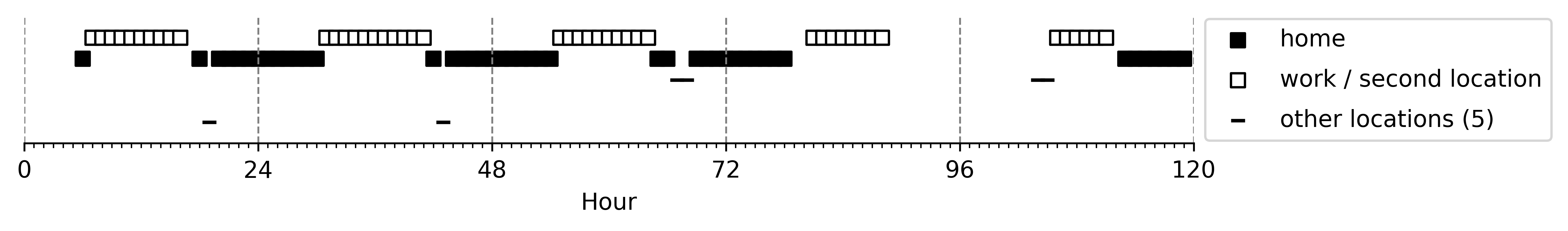}
         \caption{ }
         \label{fig:real_tvs}
     \end{subfigure}
     \par\bigskip
     \hspace*{-0.65cm}
     \begin{subfigure}[b]{0.55\textwidth}
         \centering
         \includegraphics[width=\textwidth]{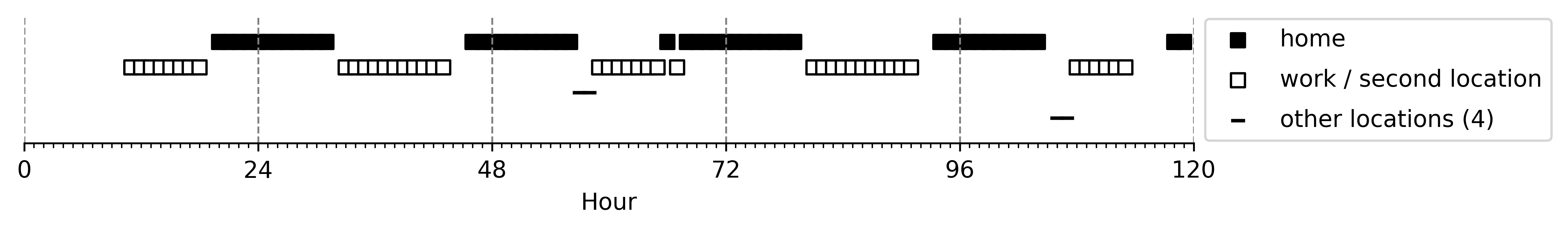}
     \end{subfigure}
     \hspace*{-0.65cm}
     \begin{subfigure}[b]{0.55\textwidth}
         \centering
         \includegraphics[width=\textwidth]{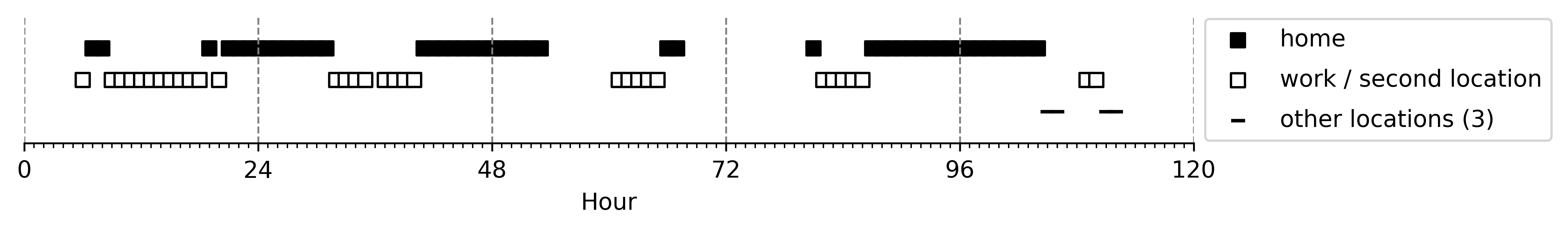}
     \end{subfigure}
     \hspace*{-0.65cm}
     \begin{subfigure}[b]{0.55\textwidth}
         \centering
         \includegraphics[width=\textwidth]{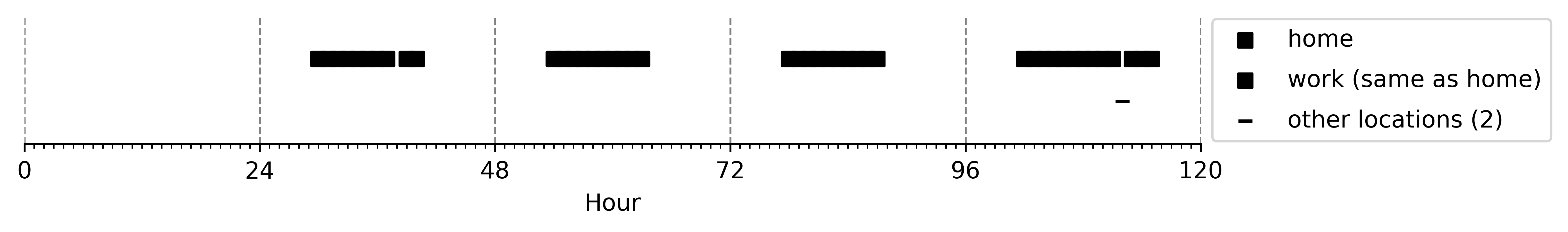}
     \end{subfigure}
     \hspace*{-0.65cm}
     \begin{subfigure}[b]{0.55\textwidth}
         \includegraphics[width=\textwidth]{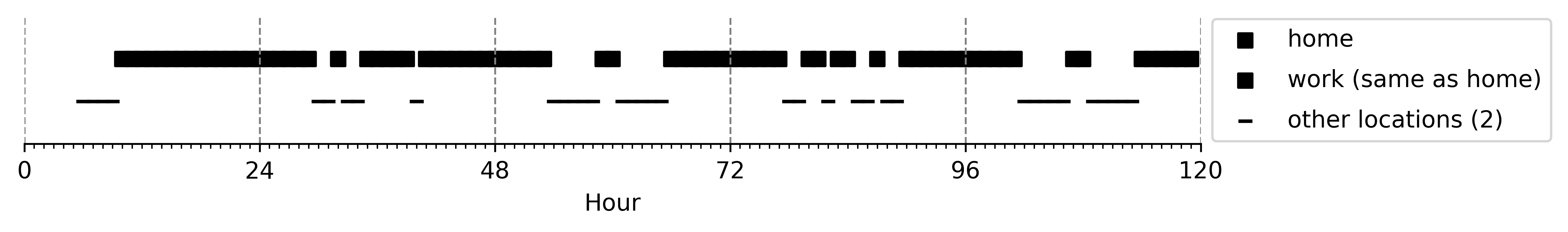}
         \caption{ }
         \label{fig:syn_tvs}
     \end{subfigure}
        \caption{
Mobility patterns. (a) Real data samples. (b) Synthetic data samples.
Each plot illustrates a user's "stay trajectory": their sequence of locations over a 5-day workweek.  Hour is on the x-axis. A point is plotted above each hour of reported data, representing where the user spent the most time within the hour.
The y-axis indicates total time the user spent in a location relative to their other locations.
}
        \label{fig:tvs}
\end{figure}

We visualize mobility patterns at the individual level by plotting stay trajectories and compare plots between the real and synthetic samples. 
These are shown in Figure \ref{fig:tvs}.\footnote{A larger set of stay trajectory plots can be viewed at: 
\url{https://github.com/aberke/lbs-data/blob/master/trajectory_synthesis/evaluation/final_eval_plots.ipynb}
}
Each plot represents a user's stay trajectory with visited locations over their 1-hour intervals. They illustrate variations in user tendencies to return to home and work locations or explore new locations.

\end{document}